\documentclass{article}
\usepackage{spconf,amsmath,graphicx}
\usepackage{comment}
\usepackage{graphicx}
\usepackage{color}
\usepackage{amsfonts}
\usepackage{wrapfig}
\usepackage{capt-of}

\usepackage{xcolor}
\definecolor{airforceblue}{rgb}{0.36, 0.54, 0.66}

\title{High-Fidelity Face Swapping with Style Blending}

\name{Xinyu Yang ~~~ Hongbo Bo}
\address{x.yang28@lancaster.ac.uk}

\begin{document}
\ninept
\maketitle
\begin{abstract}

Face swapping has gained significant traction, driven by the plethora of human face synthesis facilitated by deep learning methods.
However, previous face swapping methods that used generative adversarial networks (GANs) as backbones have faced challenges such as inconsistency in blending, distortions, artifacts, and issues with training stability.
To address these limitations, we propose an innovative end-to-end framework for high-fidelity face swapping. First, we introduce a StyleGAN-based facial attributes encoder that extracts essential features from faces and inverts them into a latent style code, encapsulating indispensable facial attributes for successful face swapping.
Second, we introduce an attention-based style blending module to effectively transfer Face IDs from source to target. To ensure accurate and quality transferring, a series of constraint measures including contrastive face ID learning, facial landmark alignment, and dual swap consistency is implemented.
Finally, the blended style code is translated back to the image space via the style decoder, which is of high training stability and generative capability. Extensive experiments on the CelebA-HQ dataset highlight the superior visual quality of generated images from our face-swapping methodology (as shown in Fig.~\ref{fig:beamer}) when compared to other state-of-the-art methods, and the effectiveness of each proposed module. Source code and weights will be publicly available.

\end{abstract}
\section{Introduction}
Face swapping is an emerging yet controversial topic in face synthesis. It holds significant potential for positive applications in entertainment, film and television production, human-computer interaction, and privacy protection \textit{etc.}.
This task involves transferring the identity of a source face image to the target image while retaining the illumination, head posture, facial expression, background, and other attribute information of the target image. It poses a significant challenge because there is no available ground truth for the swapped image, and objective evaluation metrics are scarce as most assessments rely on subjective perception.

While early face swapping methods required numerous images of the source identities and the target identities for model training~\cite{bitouk2008face,korshunova2017fast}, the trained models were specifically tailored to these identities,  limiting transferability to unseen identities.
Leveraging the powerful generative capacity of GAN, recent face-swapping methods have achieved high transferability on unseen identities, generating images effectively blending source and target identities along with facial attribute information \cite{zhu2021one,li2019faceshifter,xu2021facecontroller,cui2023face}. However, these methods struggle to balance between the identity similarity and the retention of target's low-level details, such as illumination and background, leading to frequent artifacts and low-fidelity generated contents.

The development of image generation methods \cite{karras2019style,karras2017progressive,choi2018stargan,zhu2017unpaired} has greatly advanced the face synthesis task. Most notably, StyleGAN \cite{karras2019style} introduces the concept of style code, which enhances the controllability of the synthesis process and enables the generation of high-resolution photorealistic faces.
The style code in the potential latent space $\mathcal{W}$ is projected from a random vector in the initial latent space $\mathcal{Z}$. It has been observed that the projected style code is abundant with relevant facial attributes, playing a crucial role in facilitating the generation of visually compelling facial images.
Inspired by this controllable face synthesis technique, we introduce a novel face-swapping pipeline that can leverage the style code for synthesizing faces in the source image.
Our approach prioritizes generating high-quality faces while also balancing identity similarity as well as the background details in the target image.

Our approach consists of three key components: facial attributes encoder, style blending module, and style decoder.
Our facial attributes encoder first extract both source and target face features that encapsulate the facial attributes and identity information, and then mapping them to a latent space at multiple levels of scales (feature pyramids).
The target embeddings are reused as the decoder's input, which is crucial for retaining the low-level details of the original targets in the generated images.
Secondly, the source embeddings and the target embeddings are blended using a style blending module (SBM) that utilizes multi-head cross attention (MHCA). We propose contrastive face ID learning, facial landmark alignment, and dual swap consistency to guide MHCA to pay attention only to the face attributes from the target embeddings and only the facial identities from source embeddings.
Finally, with the pre-trained style decoder, the blended embeddings in the latent space are decoded into images by leveraging the low-level features that are directly fed to the decoder. This can help ensure that the generated content retains as much low-level detail as possible.

\begin{figure}[t]
\centering
\includegraphics[width=.9\linewidth]{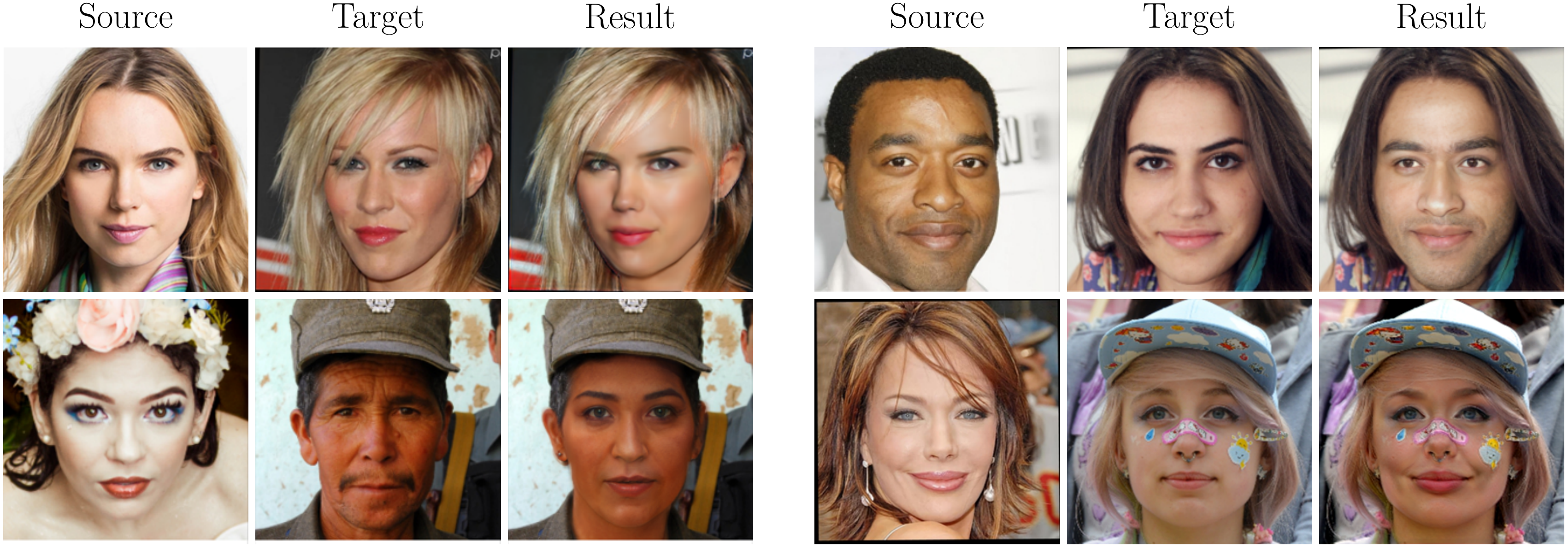}
\caption{
\footnotesize{
\textbf{Face Swap Visualisation.}
Replace the face in the target image with the face from the source. Our results are displayed on the right.
}}
\label{fig:beamer}
\end{figure}

\section{Face Swapping with Style Blending}
\begin{figure*}[t]
	\centering
	\includegraphics[width=\linewidth]{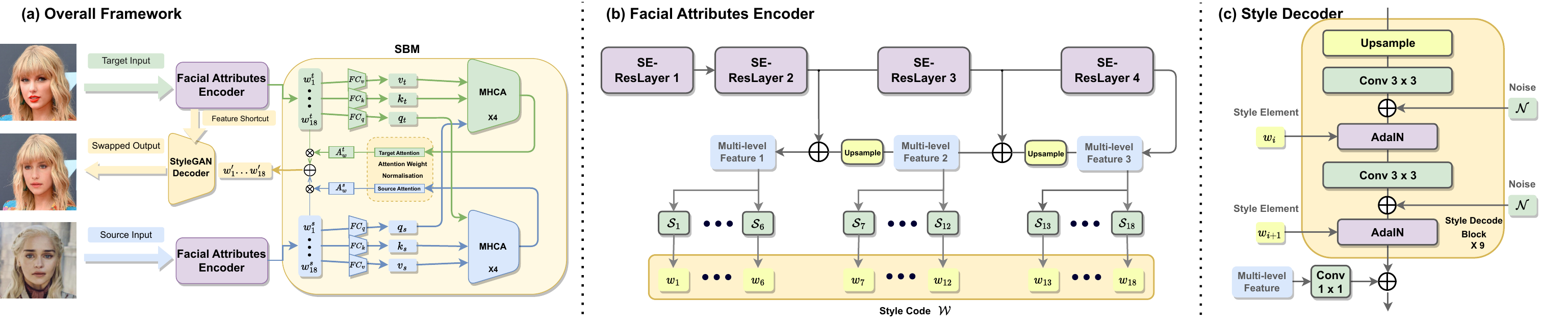}
	\caption{\footnotesize{
			\textbf{Overview of Face Swapping system.}
			(a) Overall framework; (b) Facial attribute encoder; (c) Style decoder.
The pipeline adheres to the encoding-blending-decoding pattern. In the encoding phase, a facial attribute encoder is utilized on both target and source images. In the blending phase, the Style Blending Module (SBM) combines the target style code and source style code using a learnable attention mechanism. In the decoding, Gaussian noises are introduced to isolate high-level semantics from stochastic variation, and target features are recycled to enhance the retention of low-level details in the target.
}}
\vspace{-2mm}
\label{fig:concept}
\end{figure*}

In this section, we introduce an end-to-end framework for high-fidelity face swapping. Our framework follows the traditional GAN methods which consists of a generator ${\rm G}$  and a discriminator ${\rm V}$.
In our generator, the facial attributes encoder ${\rm E}$ encodes both the target input image $\mathcal{I}^t$ and the source input image $\mathcal{I}^s$ which randomly sampled from dataset into the style latent space $\mathcal{W}$, denote as $\mathcal{W}^t$ and $\mathcal{W}^s$ respectively. In accordance with the style code design of StyleGAN \cite{karras2019style}, this study employs 18 style elements, denoted as $w_i$ for each element, each consisting of 512 dimensions, resulting in $\mathcal{W} \in \mathbb{R}^{18\times512}$. A style blending module ${\rm B}$, is utilized to seamlessly integrate $\mathcal{W}^s$ and $\mathcal{W}^t$, into a new swapped style code $\mathcal{W}^\prime$. Finally, the pretrained style decoder ${\rm D}$, is used to translate $\mathcal{W}^\prime$ back into the image $\hat{\mathcal{I}}^{s \to t}$. The face swapping generator ${\rm G}$ and the details of the style blending module (SBM) are graphically depicted in Fig. \ref{fig:concept}.

\subsection{Facial Attributes Encoder}

The structure of the facial attributes encoder is illustrated in Fig.~\ref{fig:concept}b. The encoder is composed of a SE-ResNet50 backbone \cite{hu2018squeeze}, along with a feature pyramid structure \cite{lin2017feature} and multiple style heads. The SE-ResNet50 backbone leverages the Squeeze-and-Excitation (SE) module to extract feature representations of the input image. This module adapts channel-wise feature response \cite{hu2018squeeze}, allowing for more stable training and the separation of style elements. Meanwhile, the feature pyramid structure enables the exploration of facial attributes at different levels, with each level responsible for extracting different style elements. In the decoder phase, these target multi-level features are reused to retain low-level details.

The feature representations from the backbone are projected into a style code that contains 18 style elements $w_i$. We use three levels of feature representation, each of which corresponds to the extraction of 6 style elements. Let $\mathcal{S}_i$, where $i=(1,2,~\cdots~18)$, denote one of 18 style heads operated on the multi-level feature representation for the generation of the style elements. Each style head $\mathcal{S}_i$ consists of several \texttt{Conv} with \texttt{LeakyReLU} layers (depending on the input resolution) and a final \texttt{FC} layer for projecting to a unified $\mathbb{R}^{512}$ latent space. The latent code is the concatenation of the style elements $\mathcal{W}^t=\mathbin\Vert^{18}_{i=1}w_i^t$, $\mathcal{W}^s=\mathbin\Vert^{18}_{i=1}w_i^s$ for both the target style code and the source style code respectively.

\subsection{Style Blending Module}
The style blending module (SBM) is introduced to blend source facial attributes with the target context. Operated in the style latent space $\mathcal{W}$, SBM takes both source style code $\mathcal{W}^s$ and the target style code $\mathcal{W}^t$ as input. Leveraging the cross-attention, it adaptively blends the target style elements into the source style elements. As shown in Fig.~\ref{fig:concept}, the style code $\mathcal{W}^s$ and $\mathcal{W}^t$ are encoded to the values $v_s$, $v_t$, the keys $k_s$, $k_t$ and the queries $q_s$, $q_t$ with three different linear transformation functions (\texttt{FC} layers) respectively.
The multi-head cross-attention (MHCA) operations are then applied for the purpose of mixing features using learnable attention weights:
\begin{align}\label{equ:}
	\small
		A^t= {\rm softmax}	\Big(\frac{q_sk_t}{\sqrt{d_{k_t}}}\Big)v_t~, ~
		A^s= {\rm softmax}	\Big(\frac{q_tk_s}{\sqrt{d_{k_s}}}\Big)v_s~,
\end{align}
where $A^s$ denotes the source attention and $A^t$ represents the target attention. Our SBM contains four of those MHCA-transformer layers. After that, the attentions for each style element are normalised and concatenated (denoted as $\mathbin\Vert$) respectively by
\begin{equation}\label{equ:}
	\small
	{
		{\rm A}_{w}^t=\mathbin\Vert^{18}_{i=1}\frac{e^{A_{i}^t}}{e^{A_{i}^s}+e^{A_{i}^t}} ~,\quad
		{\rm A}_{w}^s=\mathbin\Vert^{18}_{i=1}\frac{e^{A_{i}^s}}{e^{A_{i}^s}+e^{A_{i}^t}} ~.
	}
\end{equation}
Finally, $A_w^t$ and $A_w^s$ are used to merge with the target and source style codes as the final swapped code $\mathcal{W}^\prime$ by
\begin{equation}\label{equ:}
{
		\mathcal{W}^\prime=A_w^t\otimes\mathcal{W}^t\oplus A_w^s\otimes\mathcal{W}^s ~,
	}
\end{equation}
where $\otimes$ denotes the element-wise broadcast multiplication and $\oplus$ denotes the element-wise broadcast addition.

\subsection{High-Fidelity Face Generation}
The style decoder projects the style code in latent space back to the image space.
As illustrated in Fig.~\ref{fig:concept}c, it contains 9 layers of style decode blocks. Each layer is responsible for synthesizing images from low to high resolutions (from $4^2$ to $1024^2$). Thus, it progressively adds the details and improves qualities along with the increase of the resolution.

The style element is injected via the adaptive instance normalization AdaIN \cite{huang2017arbitrary} layers after each \texttt{Conv} layer. Let's suppose the input feature to the AdaIN layer is $x$ and the injection style element is $w_i$, the AdaIN operation is defined as
\begin{equation}\label{equ:}
	\footnotesize{
		{\rm AdaIN}(x, w_i)= \frac{x-\mathbb{E}[x]}{\mathbb{V}[x]+\epsilon} * \gamma_{w_i} +\beta_{w_i} ~,
	}
\end{equation}
where $\mathbb{E}[x]$ is the instance mean for each channel, $\mathbb{V}[x]$ is the instance standard deviation for each channel. $\epsilon$ is a value added to the denominator for numerical stability. $\gamma_{w_i}$ and $\beta_{w_i}$ are the decomposition of the affine transformation of style element $w_i$: $[\gamma_{w_i}, \beta_{w_i}] = {\mathrm T}w_i$, where $\mathrm{T}$ is the learnable affine transformation matrix.

Following the StyleGAN \cite{karras2019style}, Gaussian noises $\mathcal{N}$ are injected in the pipeline for a better separation of the high-level semantics from stochastic variation. In order to recover as much of the low-level details of the target image as possible, we build a short pathway for the target image feature added after the style decode block.

\subsection{Training Objectives}

The objective of the face-swapping task is to transfer the facial biometrics of a source image onto a target image while retaining other pertinent low-level visual details.
The resulting image should meet three distinctive requirements: i) it should be photorealistic and free of any artifacts; ii) it should maintain the identity of the source input; iii) it should exhibit similarity in facial expression, head pose, background, and illumination to the target input. To achieve these objectives, this section introduces a series of regularizations during the model learning process.

\vspace{5pt}
\noindent\textbf{Adversarial Learning}.
To ensure that the generated contents are realistic, we apply adversarial training for our face-swapping generator {\rm G}. Let $\mathcal{L}_{adv}$ denote the adversarial objective for restricting the generated image $\hat{\mathcal{I}}_{s\to t}$ to be realistic. A discriminator {\rm V} which is implemented via a multi-scale classifier \cite{park2019semantic} on the downsampled output images is employed to quantify how realistic $\hat{\mathcal{I}}_{s\to t}$ is. The training of the generator {\rm G} and the discriminator {\rm V} follows the adversarial learning paradigms where a hinge loss \cite{lim2017geometric} is applied on the discriminator training process:
\vspace{-2mm}
\begin{align}
    \mathcal{L}^{\rm G}_{adv} = &-\mathbb{E}_{\mathcal{I}^s,~ \mathcal{I}^t \sim \mathfrak{D}} {\rm V}({\rm G}(\mathcal{I}^s,\mathcal{I}^t)) ~, \label{eq:Gloss} \\
    \mathcal{L}^{\rm V}_{adv}  = &-\mathbb{E}_{\mathcal{I} \sim \mathfrak{D}}\big[\min (0,-1+{\rm V}(\mathcal{I}))\big] \\
    &-\mathbb{E}_{\mathcal{I}^s,~ \mathcal{I}^t \sim \mathfrak{D}}\big[\min(0,-1-{\rm V}({\rm G}(\mathcal{I}^s,\mathcal{I}^t)))\big] ~, \label{eq:Vloss}
\end{align}
where target image $\mathcal{I}^t$, source image $\mathcal{I}^s$ and a random image $\mathcal{I}$ are sampled from a face dataset $\mathfrak{D}$.

\vspace{5pt}
\noindent\textbf{Facial Identity Learning}.
To preserve the identity of the source image as much as possible while keeping the generated content realistic, we ensure consistency of the identity feature between the generated image and the original source image:
\begin{align}
		\mathcal{L}_{id} & = 1-\cos \left(\mathcal{F}_{id}\left(\hat{\mathcal{I}}_{s\to t}\right), \mathcal{F}_{i d}\left(\mathcal{I}^{s}\right)\right)~,
\end{align}
where $\mathcal{F}_{id}$ is the ID extractor implemented with the pretrained ArcFace \cite{deng2019arcface} model; $\cos$ represents the cosine similarity.

\begin{figure}[t]
	\centering
	\includegraphics[width=\linewidth]{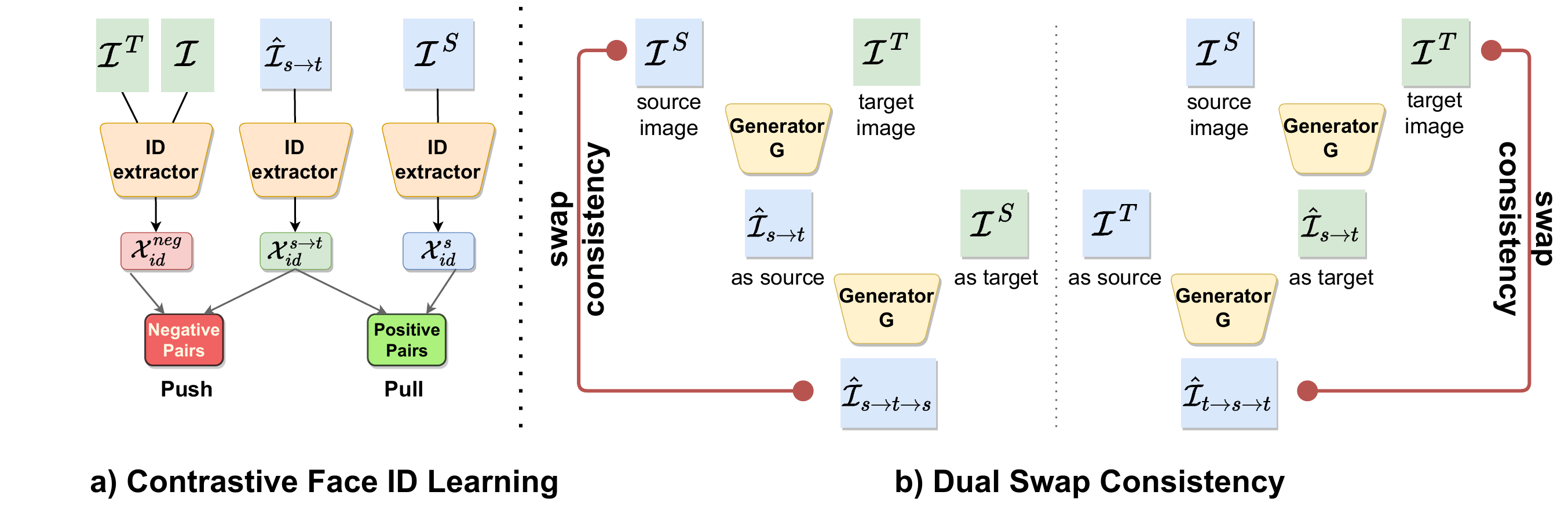}
	\caption{\footnotesize{
			\textbf{Contrastive Face ID Learning and Dual Swap Consistency}.
{a) illustrates the contrastive face ID learning. This loss pushes negative pairs away and pulls positive pairs closer. b) depicts the swap consistency loss. This loss expects the generator can revert back to the original input after applied twice. Two cases of swap consistency are presented.
			}
}}
\label{fig:loss}
\vspace{-5mm}
\end{figure}

\vspace{5pt}
\noindent\textbf{Contrastive Face ID Learning}.
To mitigate the problem of dissimilarity between the generated and source faces, it is imperative to ensure that the generated face differs significantly from the target face.
We propose to contrast the generated facial image against various negative samples.
More specifically, the objective is to minimize the feature distance between the generated face and the source face while simultaneously maximizing its distance from other faces. Following the InfoNCE \cite{van2018representation}, let $S_i =\mathcal{S}_{\text{sim}}(\mathcal{X}^i_{id}, \mathcal{X}_{id}^{s\to t})/\tau$ we formulate our ID contrast as
\begin{equation}\label{equ:contrastive}
	{
		\mathcal{L}_{con}=-\log\frac{\exp(S_s)}
		{\exp(S_t)+\sum_{n\sim p_{\text{neg}}}\exp(S_n)
		} ~,
	}
\end{equation}
where $\mathcal{X}_{id}^t$, $\mathcal{X}_{id}^s$ and $\mathcal{X}_{id}^{s\to t}$ denote the target, source, prediction vectors, which are generated by a ArcFace $\mathcal{F}_{id}$ ID extractor. $\mathcal{S}_{\text{sim}}$ represents the normalized dot product operation; $P_{\text{neg}}$ denotes the negatives that sampled from other faces. $\tau$ is the InfoNCE temperature coefficient. Fig.~\ref{fig:loss}a provides a details illustration of the ID contrastive loss.

\vspace{5pt}
\noindent\textbf{Face Reconstruction}.
When the source and target images are the same input, the model should be able to reconstruction the original image. Following \cite{li2019faceshifter,xu2021facecontroller}, we define face reconstruction in our model by minimising
\begin{align}
		\mathcal{L}_{rec} & = \left\{\begin{array}{ll}
				\frac{1}{2}\left\|\hat{\mathcal{I}}^{s\to t}-\mathcal{I}^{t}\right\|_{2}^{2} & \text{if } \mathcal{I}^{t}	= \mathcal{I}^{s} \\
				0 & \text{otherwise } ~.
		\end{array}\right.
\end{align}
We occasionally force the source input and the target input to be the same image $\mathcal{I}^{t}  = \mathcal{I}^{s}$, so that the expected output should recover the input image in pixel level. This constraint is disabled ($\mathcal{L}_{rec}=0$) when the source input and the target input are different.
Following the curriculum learning strategy \cite{bengio2009curriculum}, we propose a dynamic controller $P_\pi$ to control the probability of source input and the target input being the same image. We start from easy cases where $P_\pi$=1 during the early learning stages and gradually decrease $P_\pi = 0$ along the training. Empirical evidence indicates that this strategy enhances training stability and accelerates model convergence.

\vspace{5pt}
\noindent\textbf{Facial Landmark Alignment}.
In order to maintain consistency in facial expressions and poses of the image $\mathcal{I}^{t}$ following a face swap, we utilize a regularized approach incorporating facial landmark alignment between $\mathcal{I}^t$ and $\hat{\mathcal{I}}^{s\to t}$:
\begin{align}
		\mathcal{L}_{{lm}} & = \frac{1}{2}\left\|\mathcal{F}_{lm}({\mathcal{I}^t})-\mathcal{F}_{lm}\left(\mathcal{I}^{s\to t}\right)\right\|^{2}_{2} ~,
\end{align}
where $\mathcal{F}_{lm}$ is a differentiable facial landmark extractor \cite{sun2019high} that can generate 19 facial landmark keypoints for $\mathcal{I}^t$ and $\mathcal{I}^{s\to t}$. L2 loss is applied for regularising the consistency of those key points.

\begin{figure*}
    \begin{minipage}{0.8\linewidth}
        \includegraphics[width=\linewidth]{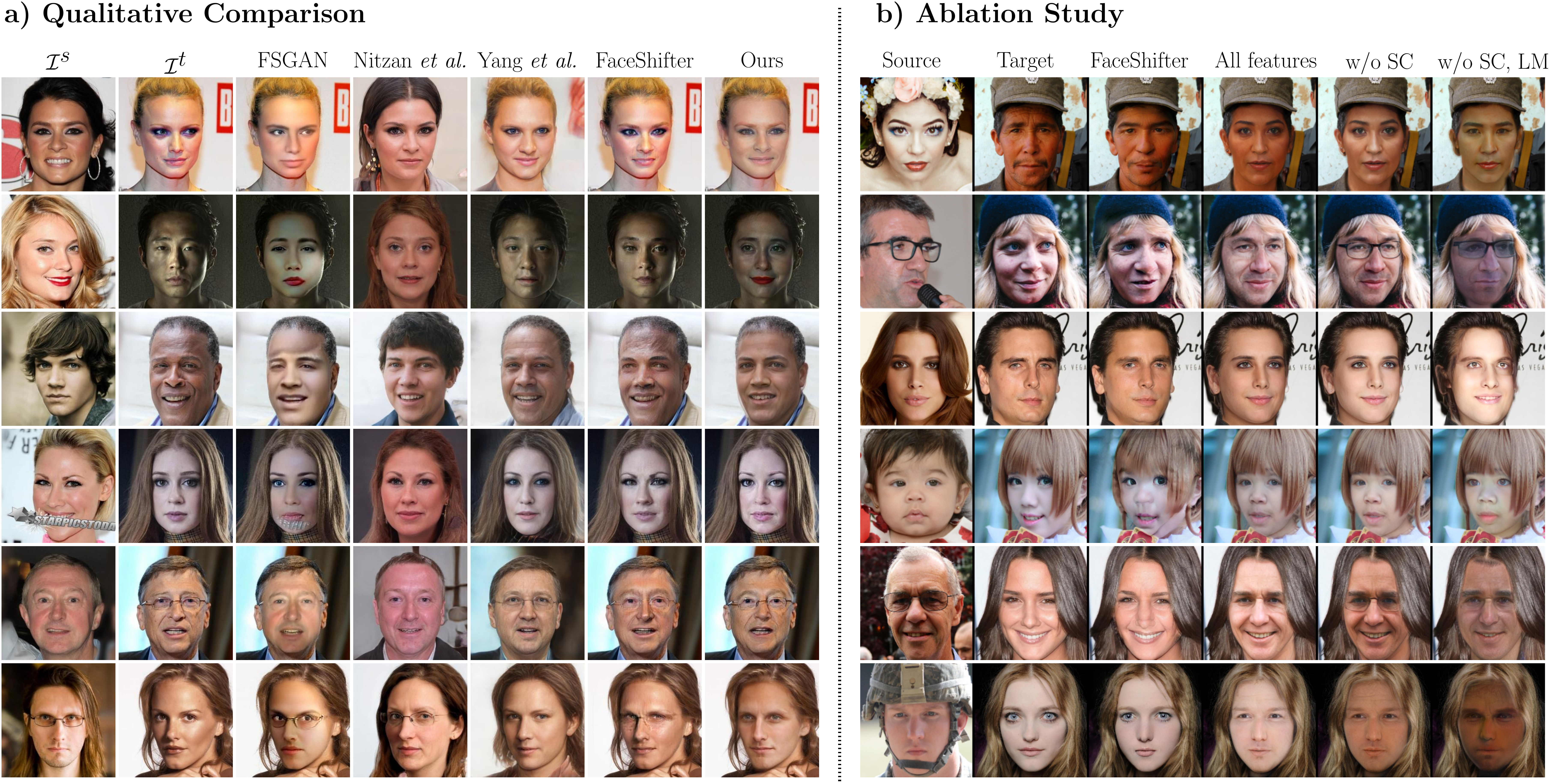}
        \caption{
		\footnotesize{
			\textbf{Qualitative Comparison with the SOTA Methods.}
			{Qualitative Comparison of our method over others on CelebA-HQ test set.
Each column, from left to right, displays the source input, target input, results from FSGAN \cite{nirkin2019fsgan}, Nitzan \textit{et al.} \cite{nitzan2020face}, Yang \textit{et al.} \cite{yang2021shapeediter}, FaceShifter \cite{li2019faceshifter}, and our method.
}}}
\vspace{-5mm}
        \label{fig:comp}
    \end{minipage}
    \hfill
    \begin{minipage}{0.18\linewidth}
        \includegraphics[scale=0.3]{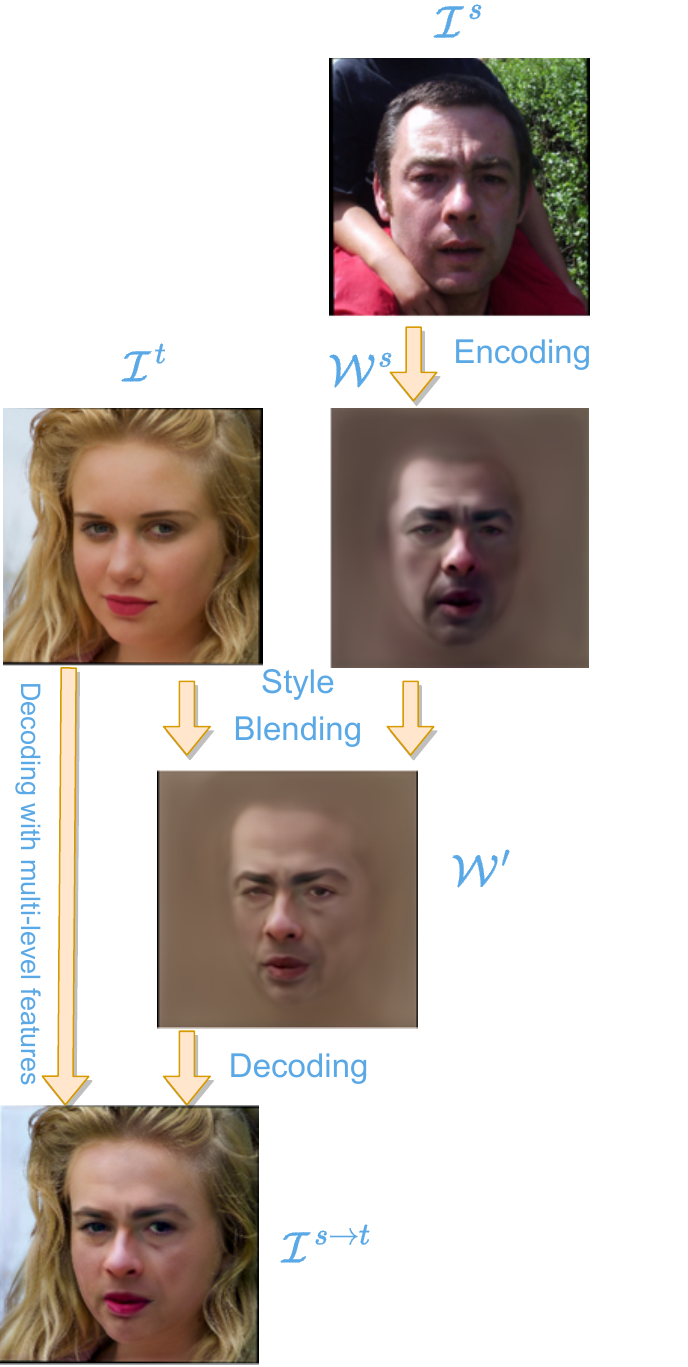}
       \caption{
		\footnotesize{
			\textbf{How It Work?}
}}
\label{fig:mech}
    \end{minipage}
\end{figure*}


\vspace{5pt}
\noindent\textbf{Dual Swap Consistency}.
The primary challenge in the task of face swapping lies in the absence of a reliable ground truth against which to evaluate the quality of the generated content.
To address this issue, we propose dual swap consistency as a means of enhancing the regularization of the generated face.Specifically, this approach involves ensuring that the generated face is capable of being swapped back to its original state after undergoing two separate passes through our generator {\rm G}.
As depicted in Fig.~\ref{fig:loss}b, there are two scenarios to consider. Firstly, when the target input is $\mathcal{I}^s$ and the source input is the first round swap output $\mathcal{I}^{s\to t}$, the model is required to reconstruct $\mathcal{I}^s$ in the subsequent round of swapping. Secondly, when the source input is $\mathcal{I}^t$ and the target input is $\mathcal{I}^{s\to t}$, the generator is expected to reconstruct $\mathcal{I}^t$. Therefore, the loss can be formulated accordingly. Let $\tilde{ \mathcal{I}^{t}}={\rm G}\left({\rm G}\left(\mathcal{I}^{s}, \mathcal{I}^{t}\right), \mathcal{I}^{s}\right)$ and $\tilde{ \mathcal{I}^{s}}={\rm G}\left(\mathcal{I}^{t}, {\rm G}\left(\mathcal{I}^{s}, \mathcal{I}^{t}\right)\right)$, this consistency can be expressed as
\begin{align}
		\mathcal{L}_{swap} & = \frac{1}{2}\left\|\tilde{ \mathcal{I}^{t}}-\mathcal{I}^{s}\right\|_{2}^{2}+\frac{1}{2}\left\|\tilde{ \mathcal{I}^{s}}-\mathcal{I}^{t}\right\|_{2}^{2} ,
\end{align}
where L2 regularisation is utilized in training to ensure the prediction is consistent.

\vspace{5pt}
\noindent\textbf{Overall Objectives}.
The overall training objectives for discriminator is $\mathcal{L}_{{\rm V}}=\mathcal{L}^{{\rm V}}_{adv}$ as demonstrated in Equation \ref{eq:Vloss}. While the overall loss of the generator {\rm G} is the weighted summation of adversarial learning $\mathcal{L}_{adv}^{\rm G}$, facial identity learning $\mathcal{L}_{id}$, contrastive face ID learning $\mathcal{L}_{con}$, face reconstruction $\mathcal{L}_{rec}$, facial landmark alignment $\mathcal{L}_{lm}$ and dual swap consistency $\mathcal{L}_{swap}$,
\begin{equation}\label{equ:all} 
	{
		\mathcal{L}_{{\rm G}}	=\mathcal{L}_{adv}^{\rm G}+\lambda_{1}\mathcal{L}_{id}+\lambda_{2}\mathcal{L}_{con}+\lambda_{3}\mathcal{L}_{rec}+\lambda_{4}\mathcal{L}_{lm}+\lambda_{5}\mathcal{L}_{swap}
	}
\end{equation}
where $\lambda_{1}, \lambda_{2}, \lambda_{3}, \lambda_{4}$ and  $\lambda_{5}$ are the weights for balancing the losses.

\section{Experiments and Results}

\vspace{5pt}
\textbf{Datasets.}
CelebA-HQ \cite{karras2017progressive} is a high-resolution facial dataset containing $30,000$ images at $1024^2$ resolution. It is a subset of the CelebA dataset, built for face detection, and facial landmark localisation. FFHQ \cite{karras2019style} is a dataset that contains 70,000-megapixel face images. FFHQ has a large variation of gender, age, and ethnicity.

\vspace{5pt}
\noindent\textbf{Details of Implementation.} The training set is a combination of CelebA-HQ and FFHQ. Facial images from both datasets are processed by first aligning the face with the facial landmark and then resizing the resolution of the image to $256^2$. In all experiments, the learning rate for {\rm G} and {\rm V} is fixed at $4e^{-5}$ with a batch size of $32$. We set $\tau=0.07$ in Equation \ref{equ:contrastive} and $\lambda_{1}=10$, $\lambda_{2}=5$, $\lambda_{3}=1$, $\lambda_{4}=100$, $\lambda_{5}=1$ in Equation \ref{equ:all}. The StyleGAN pre-trained weights \cite{karras2019style} are used to initialize style decoder {\rm D}.
For training stability, we freeze the parameters of decoder {\rm D } while only training the encoder {\rm E } and SBM.
Following this, we unfreeze {\rm D } but freeze {\rm E } and SBM and conduct fine-tuning with a small learning rate. The whole framework is trained for 2.4M iterations in total on 8 RTX2080Ti GPUs.


\vspace{5pt}
\noindent\textbf{Qualitative Comparison.}
The qualitative analysis is presented in Fig.~\ref{fig:comp}a. Early GAN-based face-swapping works, FSGAN \cite{nirkin2019fsgan} and FaceShifter \cite{li2019faceshifter}, were highly regarded for their notable achievements at the time. However, these methods are susceptible to producing significant artifacts, as evident in R3, R4, and R6 of Fig.~\ref{fig:comp}a, thus failing to transfer face ID accurately.
Subsequent style-code-based methods like Nitzan \textit{et al.} \cite{nitzan2020face} and Yang \textit{et al.} \cite{yang2021shapeediter} showed photorealistic results with minimal artifacts. However, some low-level details such as hairstyle, background, and head pose from the target image were not maintained, as seen in R1, R3, R4, and R5 in Fig.~\ref{fig:comp}a.
Our approach, on the other hand, effectively addresses the challenges of achieving a balance between face ID similarity and preserving details (see the last column in Fig.~\ref{fig:comp}a). Our method produces the most realistic face-swapping effects.

\vspace{5pt}
\noindent\textbf{Ablation Studies.}
An ablation study was carried out on two critical constraints that were utilized in training our methodology - namely, Dual Swap Consistency (DSC) and Facial Landmark Alignment (LMA). The results of the study have been illustrated in Fig.~\ref{fig:comp}b. Observation of the results reveals that the omission of the swap consistent loss during training results in the loss of low-level details such as glasses and skin color. Additionally, the absence of facial landmark constraints during model training leads to a significant decrease in visual perception, with head pose and facial expressions not being preserved.


\noindent\textbf{Further Analysis.}
We take a further study on how the face swap system works. Fig.~\ref{fig:mech} illustrates the example of face swapping in our system. We visualized the intermediate output $\mathcal{W}^s$ and $\mathcal{W}^\prime$ from the model. We observe that the face attributes encoder first learns the essential face attributes from the source image. Then, the SBM edits the face attributes adaptively by the guidance of the target style code. We can see the SBM effectively synthesizes the facial expression and the head pose \textit{etc.}. Lastly, it is rendered leveraging the multi-level features that contain low-level details from the target image.


\section{Conclusion}
This paper presents a cutting-edge approach to face-swapping that harnesses the advanced style-code-based model for generating photorealistic faces. Our unique style blending module, which operates within the latent space, showed remarkable results in achieving high-quality face synthesis while ensuring a delicate balance between maintaining the source's identity similarity and incorporating the target's fine details. This work highlights the potential of a tuned style-code-based model for efficient and successful face-swapping, bringing us closer to realizing high-resolution and photorealistic face swaps.

\clearpage
\bibliographystyle{IEEEbib}
\bibliography{main}
\end{document}